\documentclass{article}

\usepackage{arxiv}

\usepackage[utf8]{inputenc} 
\usepackage[T1]{fontenc}    
\usepackage{hyperref}       
\usepackage{url}            
\usepackage{booktabs}       
\usepackage{amsfonts}       
\usepackage{nicefrac}       
\usepackage{microtype}      
\usepackage{lipsum}
\usepackage{amsmath}
\usepackage{pgfplots}
\usepackage{subfigure}

\usepackage{hyperref}
\hypersetup{
    colorlinks=true,
    linkcolor=blue,
    urlcolor=red,
    pdftitle={Research Statement}
    }
\usepackage{fancyhdr}
\usepackage{setspace}
\usepackage{amsmath}
\usepackage{amssymb}
\usepackage{makeidx}
\usepackage{lmodern}
\usepackage{fourier}
\title{Feasibility Study on Active Learning of Smart Surrogates for Scientific Simulations}

\author{
  Pradeep Bajracharya \\
  Rochester Institute of Technology, \\
  Rochester, New York, USA \\
  \texttt{pb8294@rit.edu} \\
  \And
    J. Quetzalc\'oatl Toledo-Mar\'in \\
  TRIUMF, \\
  Vancouver, Canada \\
  \texttt{jtoledo@triumf.ca} \\
  \And
    Geoffrey Fox \\
 University of Virginia, \\ 
 Charlottesville, Virginia, USA\\
  \texttt{gcfexchange@gmail.com} \\
  \And
  Shantenu Jha \\
  Rutgers University, \\
  Piscataway, New Jersey, USA \\
  \texttt{shantenu.jha@rutgers.edu} \\
    \And
 Linwei Wang \\
 Rochester Institute of Technology, \\
  Rochester, New York, USA \\
  \texttt{lxwast@rit.edu} \\
}
\date{}

\begin{document}
\maketitle

\begin{abstract}
High-performance scientific simulations, important for comprehension of complex systems, encounter computational challenges especially when exploring extensive parameter spaces. There has been an increasing interest in developing deep neural networks (DNNs) as surrogate models capable of accelerating the simulations. However, existing approaches for training these DNN surrogates rely on extensive simulation data which are heuristically selected and generated with expensive computation 
-- a challenge under-explored in the literature. 
In this paper, we investigate the potential of incorporating active learning into DNN surrogate training. This allows intelligent and objective selection of training simulations, reducing the need to generate extensive simulation data as well as the dependency of the performance of DNN surrogates  on pre-defined training simulations. 
In the problem context of constructing DNN surrogates for diffusion equations with sources, 
we examine the efficacy of diversity- and uncertainty-based strategies for selecting training simulations, 
considering two 
different DNN architecture.
The results 
set the groundwork for developing the high-performance computing infrastructure for \textit{Smart Surrogates} that supports on-the-fly generation of simulation data steered by active learning strategies to potentially improve the efficiency of scientific simulations.
\end{abstract}


\keywords{Machine Learning \and Deep Learning \and Active Learning \and Diffusion Solver \and Surrogate Modeling \and Encoder-decoder}


\section{Introduction}
High-performance scientific simulations are crucial in advancing our understanding of complex systems and allow accurate modeling of phenomena ranging from molecular interactions to climate patterns and astrophysics. These simulations, enabled by high-performance computing (HPC), can attain insights that were previously impossible with only observational data, allowing researchers to explore multiple hypothetical scenarios and predict respective system behaviors.

However, despite HPC advances, a limitation arises as the computational demands of the increasingly complex simulation models surpass the system's capacity, especially when exploring extensive parameter spaces. Recent developments in machine learning and deep learning have sparked an interest in creating efficient surrogate approximations for these scientific simulations. For either certain internal components of the simulations or the entire simulations, these surrogates could drastically accelerate expensive scientific simulations by several orders of magnitude \cite{fox2019learning, jha2019understanding, fox2019learningieee}. Amidst these, deep neural network (DNN) based surrogates have emerged as powerful tools for optimizing high-performance scientific simulations in multidimensional space \cite{kasim2021building, hruska2020extensible, lee2019deepdrivemd, mardt2018vampnets, bhowmik2018deep, ton2020rapid, liao2019deepdock, clyde2020regression, sun2020surrogate, gao2021phygeonet, mustafa2019cosmogan}. These approaches address both the computational challenges associated with extensive simulations and help improve the efficiency and speed of scientific inquiry.


Despite substantial recent progress, the prevalent approach in building DNN surrogates involves training these models using an extensive simulation dataset that covers the bounds of input parameter space. However, generating these data can be computationally expensive, if not prohibitive. In addition, the training data may include redundant subsets in some regions of the parameter space while insufficient in others, where changes in such training simulation are likely to result in different optimal surrogates. 


In this paper, we investigate the potential of integrating active learning in the training of DNN surrogate models towards the long-term establishment of an HPC infrastructure of \textit{Smart Surrogates} that can intelligently select on-the-fly generation of optimal training simulations towards DNN surrogate construction. Deep active learning, generally used in the absence of extensive pre-annotated data, enables selective query of labels for 
a small amount of data to achieve DNN performance that otherwise needs to be achieved with a large amount of labeled data. 
In the setting of DNN surrogate construction for high-performance simulations, 
this will enable an intelligent exploration of the parameter space and selective execution of simulation runs, 
rather than random or uniform sampling across the parameter space of the simulation models. This addresses the resource challenges associated with generating extensive simulation data 
and reduces the dependency of the DNN surrogate on training simulations pre-defined with ad-hoc assumptions 
(and thus inconsistencies arising from the difference in assumptions). 

This potential of active learning, however, has not been systematically explored in the construction of DNN surrogates for high-performance simulations, 
leaving two critial gaps of knowledge. 
First, 
it is not clear what types of data selection strategies, 
among the representative uncertainty- and diversity-based strategies
known as \textit{acquisition functions}, 
are more effective for the construction of DNN surrogates. 
Second, while architectural choices have been considered an important topic for DNN surrogates, 
how the performance of active learning may be affected by different architectures of the DNN surrogates is not understood.

In this paper, 
we attempt to provide initial insights into the above critical gaps in the specific context of diffusion equations with sources. 
Often in science, we are interested in the rate of change of some physical quantity concerning time and how this rate of change depends on the environment. The most common way to model the process is via partial differential equations. This concept underpins a myriad of phenomena, exemplified by models such as the diffusion equation in heat conduction, Schr\"odinger's equation in quantum mechanics, the Navier-Stokes equation in fluid dynamics, and others in fields ranging from electromagnetism to finance and biology. The intricacies of these problems are often dictated by their nature and specific initial and boundary conditions, with solutions varying from simple stationary answers to complex dynamic responses. Here we focus on one such instance as a use case – the stationary solution of the diffusion equation in scenarios with sources. Despite their ubiquity across various systems, diffusion processes pose substantial computational challenges, particularly when solving steady-state and time-varying equations in environments with parameters that vary significantly. These complexities are compounded when factors like the number of sources and sinks fluctuate, increasing the computational demands of simulations \cite{toledo2021deep}. DNNs have been proposed as PDE solvers \cite{van2023efficient}. In particular, convolutional neural networks have been successful mainly due to the inherent position encoding in convolutional layers. Alterantively, U-Nets that were first proposed for image segmentation \cite{ronneberger2015u} have been used in as PDE solvers as well \cite{gupta2022towards}. 
We build on these past successes and study the effectiveness of different active learning strategies in building DNN surrogates for this problem, considering both the CNN and U-Net architectures reported in previous works. 

As a first proof-of-concept, we carried out the above investigations in an \textit{emulated} active learning setting where the selection of training simulations was conducted over a large dataset of simulation data already generated offline based on uniform sampling of the parameter space. Our findings 
suggested that uncertainty-based selection of training simulations, 
especially that based on 
predicted loss of the DNN surrogates, 
has the potential to improve the accuracy of the DNN surrogates with less training simulations. 
We further found that identifying an appropriate DNN architecture for the given scientific simulations of interest is critical to bring out the benefit of active learning of DNN surrogates. 
These results provide an important basis  for the design and developments of the \textit{Smart Surrogates} HPC infrastructure to support the on-the-fly generation of simulation data steered by active learning, towards the ultimate goal of improving the efficacy and efficiency of DNN surrogate construction for scientific simulations. 

\section{Related Works}
\textbf{Diffusion Solvers:} Diffusion solvers are integral to scientific computing, given the widespread occurrence of diffusion processes across various domains. Traditional numerical methods, such as Finite Elements and Finite Differences, have been the cornerstone in this area due to their robustness and efficiency in certain scenarios \cite{belytschko2007first, schiesser2012numerical} and, in some cases, these methods can be fairly optimized \cite{rackauckas2017differentialequations}. However, the effectiveness of these methods often hinges on specific factors such as the geometry of the problem, inherent symmetries, and the problem parameters (\textit{e.g.}, diffusivity field, decay rate field, etc.), which can limit their applicability in more complex scenarios.
In response to these limitations, DNNs have emerged as promising surrogates, offering the potential to expedite simulations traditionally handled by these numerical methods \cite{farimani2017deep, han2018solving, li2020fourier, he2020unsupervised, willard2020integrating, cai2021physics}. Despite their promise, deploying neural networks in this context is challenging. Issues such as prolonged training times, the scarcity of curated datasets, and a lack of physical principles integration can significantly impede their performance and reliability in accurately modeling diffusion processes \cite{toledo2023analyzing}. Addressing these challenges is crucial for effectively integrating neural networks in diffusion solvers, ensuring both speed and accuracy in scientific computations. 

The active learning approach examined in this study will contribute towards addresses the challenge of curating large simulation datasets for learning diffusion solvers, which have not yet been investigated to our knowledge. 

\textbf{Deep Active Learning (DAL):} Active learning is an area in machine learning that deals with incremental intelligent selection of data to query for labels to achieve high model performance with low annotation cost \cite{cohn1996active}. In the context of deep learning, this practice is referred to as Deep Active Learning (DAL), which has led to the development of various data selection strategies. The strategies can be categorized into uncertainty-based, diversity-based, and hybrid methods. Uncertainty-based methods actively select examples that the DNN is most uncertain about. Examples of such methods include estimation of DNN loss
\cite{yoo2019learning, huang2021semi}, entropy \cite{joshi2009multi, gal2016dropout}, BALD \cite{houlsby2011bayesian, kirsch2019batchbald}, margin sampling \cite{roth2006margin, scheffer2001active}, and least confidence sampling based on softmax output \cite{settles2009active}. The diversity-based method aim to find a subset of diverse samples that effectively represents the data distribution. These methods often rely on either core set selection \cite{sener2017active, geifman2017deep} or density-based clustering \cite{nguyen2004active, wang2017active} to find examples that are most representative of the distribution of the entire unlabeled pool of data. Hybrid methods combine the benefit of both uncertainty-based and diversity-based approaches \cite{wang2016cost, zhou2017fine, yang2017suggestive, ash2019deep, shui2020deep, wang2022deep, kong2022neural}.

Most of these methods are derived particularly for image classification tasks, and their benefit for constructing DNN surrogates for scientific simulation have been little studied.

\textbf{Active Learning in Surrogate Modeling:} 
Some recent works have emerged in exploring the role of active learning in DNN surrogates for scientific simulations. 
A Bayesian active learning 
was proposed in \cite{wu2023deep} 
to use the 
expected information gain 
to select training simulations when 
training a neural process as the 
surrogate model of scientific simulations, 
tested on
the reaction-diffusion, heat, and local epidemic and mobility models. 
 An active learning method is presented in \cite{pestourie2020active} to reduce the number of simulations for an NN surrogate model of optical-surface components for photonic device modeling. They suggests an ensemble training-based acquisition function which selects the examples based on the error measure. 
In \cite{lye2021iterative}, 
a DNN surrogate is built to link the PDE parameter to the observables in the context of solving for PDE constrained optimizations, where the 
acquisition function is based on the optimization objective function using the DNN surrogates and selects 'k' parameters below the tolerance set. 
A two network (classifier and regressor) physics-informed active learning system, 
named ADEPT (Active Deep Ensembles for Plasma Turbulence),
is presented in 
\cite{zanisi2023efficient}.
This framework is designed to train a surrogate model for gyrokinetic turbulence, with the primary goal of minimizing the required data volume. The classifier screen identifies potential candidates from unlabelled pool and regressor's epistemic uncertainty is used to strategically select the examples for simulation.

The active learning methods considered in these works are specifically derived for the particular applications and tied to specific choices of DNN architectures;  
in the latter two cases, 
the design of acquisition functions 
are further tied to specific optimization objectives using the scientific simulations. 
Till now, 
there lacks a systematic study on the effect of different data selection strategies on DNN surrogate construction, nor how they are impacted by the choice of DNN architectures.

\section{Background: Diffusion Solver}

 Diffusion processes, prevalent in a myriad of systems, are crucial in modeling biological phenomena but bring along substantial computational challenges. Addressing both steady-state and dynamic diffusion equations becomes particularly complex in biological contexts, where environmental parameters can vary greatly. This complexity is further heightened in simulations that involve fluctuating elements like varying numbers of sources and sinks, a common scenario in biological systems. These factors significantly increase the computational load, underscoring the need for efficient and robust diffusion solvers. In biological systems, diffusion is often key to understanding processes such as nutrient transport, cellular signaling, and tissue growth, making the ability to accurately simulate these processes essential for advancing our understanding of complex biological mechanisms. 


\begin{figure}[!t]
  \centering
\includegraphics[width=6.0in]{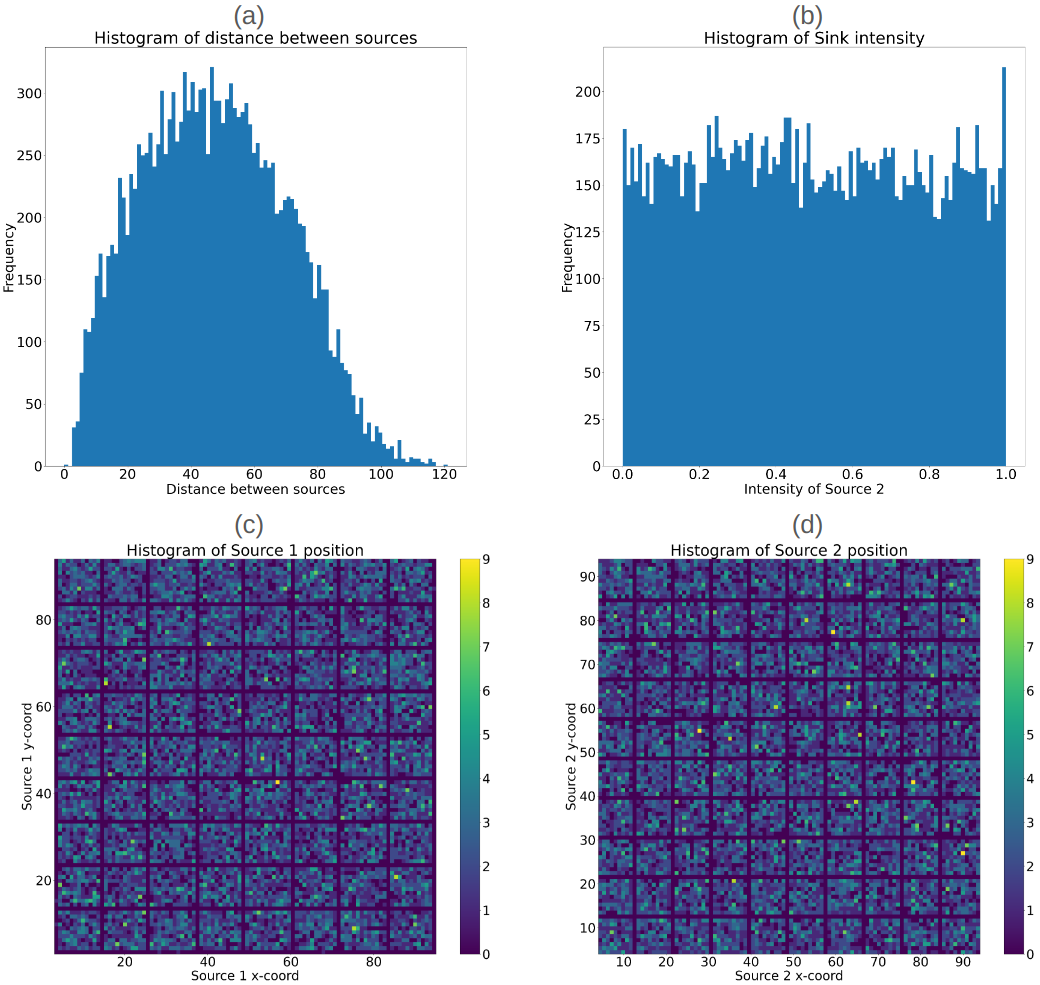}
  \caption{Distribution of parameters used to generate simulation data.  (a) Histogram of distance between two sources on the lattice. (b) Histogram of intensity of source 2 which has randomly assigned value between 0 and 1. (c-d): Histograms of the positions of (c) source 1 and (d) source 2 on the lattice.}
  \label{fig_parameter}
\end{figure}

\begin{figure}[!t]
  \centering
  \includegraphics[width=\linewidth]{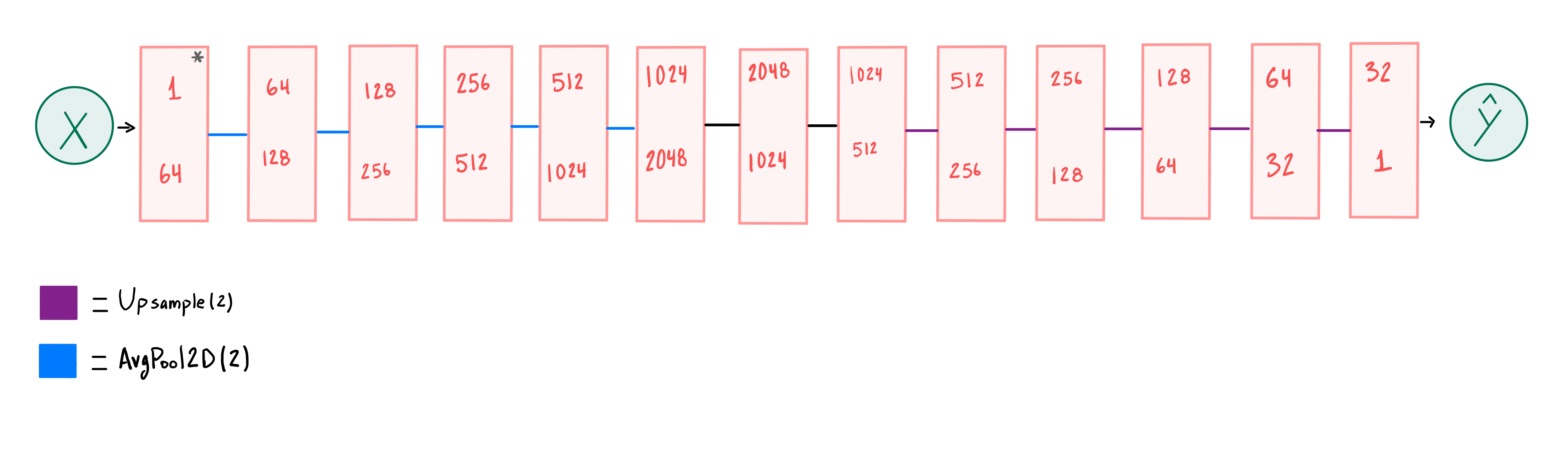}
  \vspace{-0.1cm}
  \includegraphics[width=0.36\linewidth]{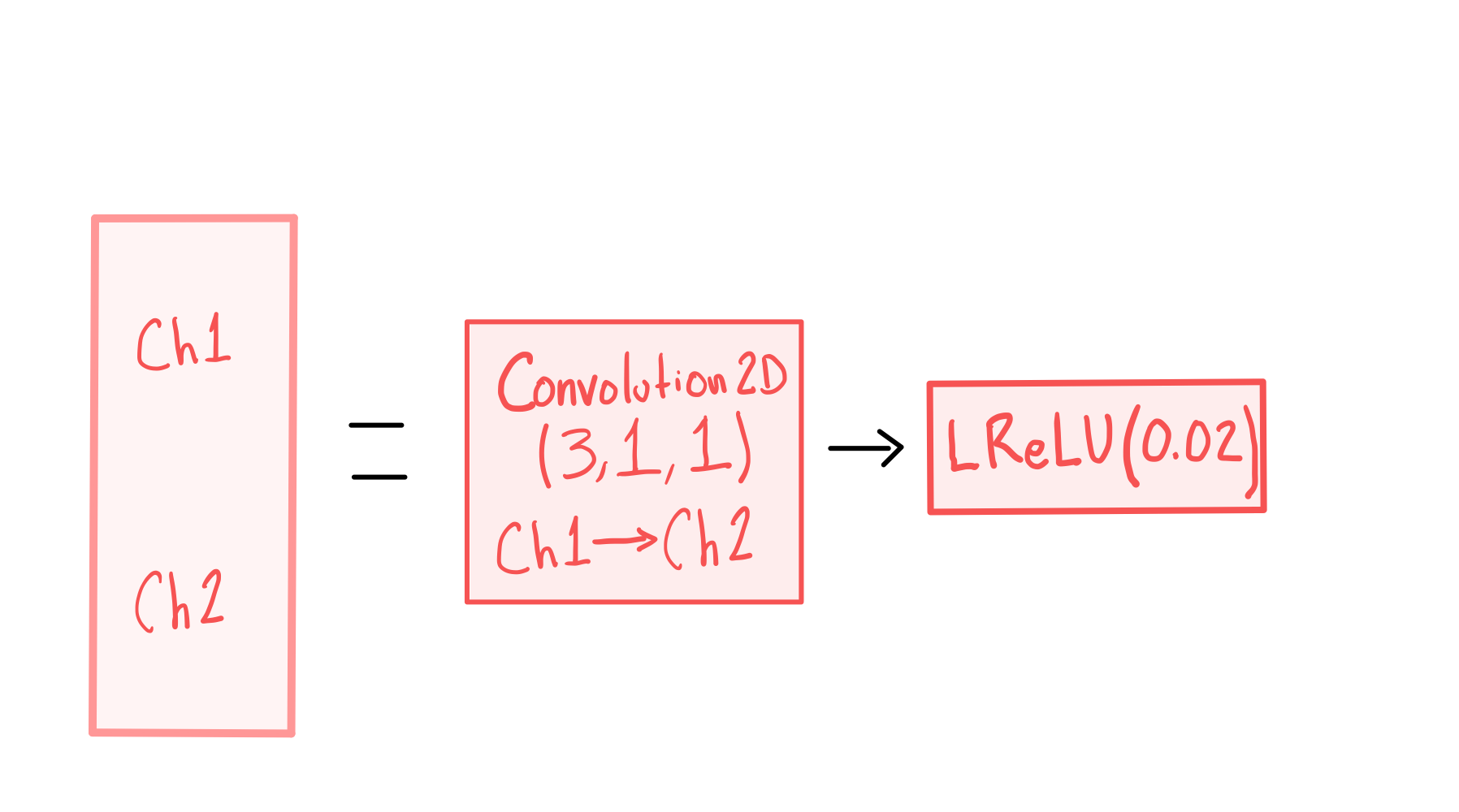}
  \caption{Deep convolutional neural network architecture for the diffusion surrogate. All convolution layers leave the input with the same height and width. Each block is composed of a convolution which increases the channels from $Ch1$ to $Ch2$, followed by a LeakyReLU with slope set at $0.02$.}
  \label{fig_CNN}
  
  \includegraphics[width=0.9\linewidth]
  {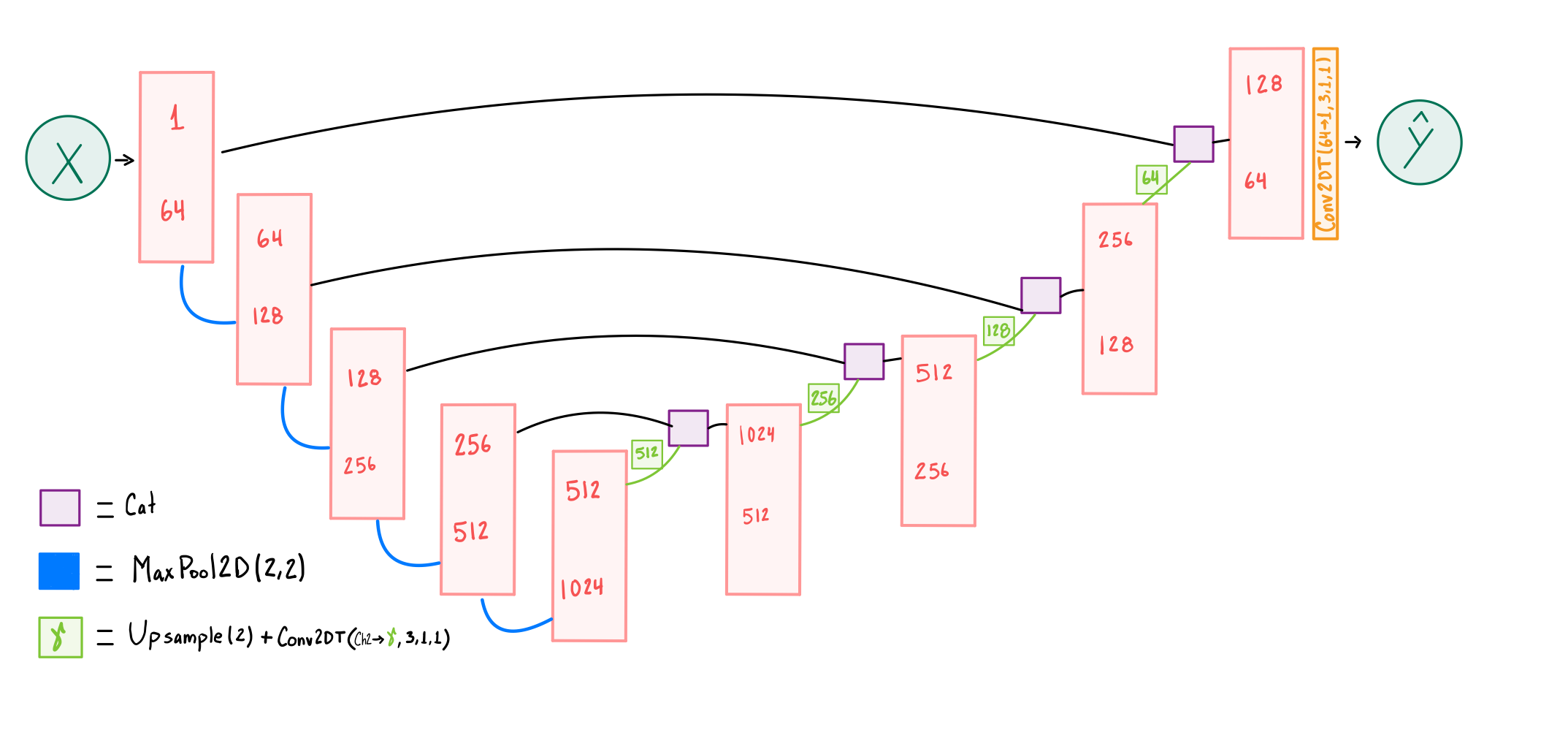}
  \includegraphics[width=0.6\linewidth]{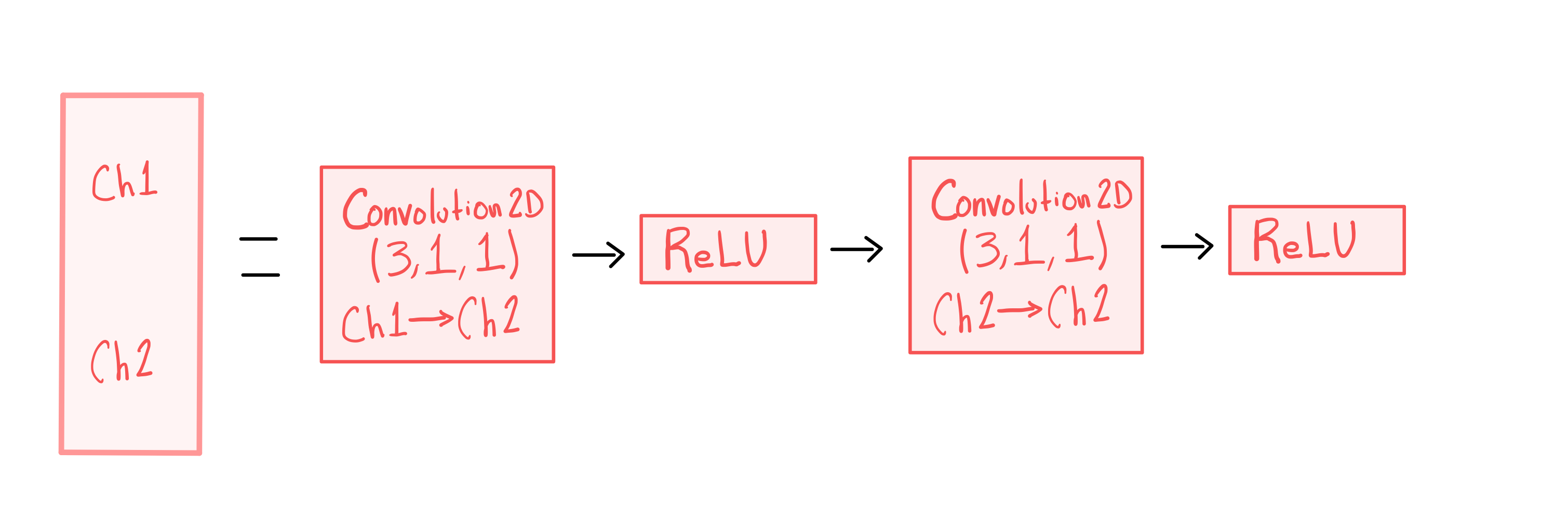}
  \caption{U-Net architecture for the diffusion surrogate. All convolution layers leave the input with the same height and width. Each block comprises a convolution that increases the channels from $Ch1$ to $Ch2$, followed by a ReLU, a convolution that leaves the dimensions unchanged, and a ReLU afterward. The main feature in U-Nets is the concatenation between blocks in the encoder and in the decoder.}
  \label{fig_UNet}
\end{figure}

\subsection{Data Generation} \label{sub_data}

The data used in this paper is motivated by biological multiscale modeling \cite{toledo2021deep}, and correspond to two sources placed in a lattice with a decay rate which acts as a sink.
To generate representative initial conditions and corresponding steady-state diffusion fields from a two-source system, we consider a 100x100 $unit^2$ lattice. In total, 20k initial configurations were created with two sources randomly positioned in the lattice. Each source, characterized by a 5-unit radius, has a constant flux of 1 for one source, while the other sources have a randomly assigned constant flux value between 0 and 1, randomly assigned using a uniform distribution. The remainder of the lattice has a field value of zero. The stationary solution to the diffusion equation with absorbing boundary conditions for each initial condition is computed using the Differential Equation package in Julia \cite{rackauckas2017differentialequations}. 
We denote $x$ as the input image that represents the initial condition layout of the two source cells, and $y$ represents the output image, which is the predicted stationary solution of the diffusion equation.

The diffusion constant is set to $D\ =\ 1\ units^2/s$ and the decay rate $\gamma\ =\ 1/400s^{-1}$, resulting in a diffusion length $l_D = \sqrt{D/\gamma}\ =\ 20units$.  This length diminishes as $\gamma$ increases and as $D$ decreases. A decrease in this length corresponds to a reduction in the field gradient as explained in \cite{tikhonov2013equations}. 
Figure \ref{fig_parameter} summarizes the distribution of some of the key parameters, 
including the distance between the two sources (a), the intensity of one source (b), 
and the positions of the two sources (c-d). 
The code for the generation of data and the data can be found \cite{jtoledo_ghb, toledo2021deep}. 

\subsection{DNN Surrogates} \label{sub_model}
 We define two DNN architectures, namely, a CNN based autoencoder and an U-Net-based structure as shown in Figure~\ref{fig_CNN} and ~\ref{fig_UNet}, respectively. 

The autoencoder's first six layers perform an average-pool operation that reduces height and width in half after each layer following the sequence \{$100^2$, $50^2$, $25^2$, $12^2$, $6^2$, $3^2$, $1^2$\} while adding channels after each layer following the sequence \{1, 64, 128, 256, 512, 1024, 2048\}. Similar to \cite{toledo2021deep}, the next six layers reduce the number of channels while increasing the height and width in the opposite order from those described above.

The first four layers of the U-Net perform max-pool operation, which, similar to CNN autoencoder, reduces the height and width to half following the sequence \{$100^2$, $50^2$, $25^2$, $12^2$ and  $6^2$\} with corresponding channels of \{1, 64, 128, 256 and 512 \} until the bottleneck is reached, where the size is $3^2$ with $1024$ channels. The subsequent four layers reverse this process, upscaling the image starting from the bottleneck and concatenating images from the downscaling side to produce features with sizes and channels opposite to those described earlier. A final layer adjusts the number of channels from $64$ to $1$ and the image size to $100^2$.


\subsection{Loss Definition} \label{sub_loss_def}
One of the sources in the input images has an intensity of 1, while the other has a uniformly distributed random number. The two sources occupy 2\% of the total image space, with the rest of the space with intensity 0. Due to the disproportionate distribution of source and space intensity, the source pixels need to be weighed to avoid high field values to wash out. The loss is thus weighed with an exponential weight and modulated with a scalar hyperparameter $w$.
\begin{equation} \label{eqn_loss}
    L^\alpha_{i\beta} = \exp\left(-\frac{(\langle i|1\rangle - \langle i|y_{\beta}\rangle)}{w}\right) \cdot (\langle i|\hat{y}_{\beta}\rangle - \langle i|y_{\beta}\rangle)^\alpha
\end{equation}

where $\alpha$ is set to 1 and $\beta$ represents the input and target tuple. 
We refer to this as weighted mean absolute error (MAE) in the remainder of this paper. 
Here $\langle|\rangle$ denotes the inner product and $|i\rangle$ is a unitary vector with the same size as $|y_{\beta}\rangle$ with all components equal to zero except the element in position $i$ which is equal to one. $|1\rangle$ is a vector with all components equal to 1 and size equal to $|y_{\beta}\rangle$. Then $\langle i|y_{beta}\rangle$ is a scalar corresponding to the pixel value at the $i^{th}$ position in $|y_{\beta}\rangle$, whereas $\langle i|1\rangle\ =\ 1$ for all $i$. Notice that high and low pixel values will have an exponential weight $\approx\ 1$ and $\approx\ exp(-1/w)$, respectively. This implies that the error associated with high pixels will have a larger value than that for low pixels. The loss function $L(\alpha)$ is the mean value over all pixels $(i)$ and a given data set $(\beta).$


\section{Methods: Active Learning of DNN Surrogates for Diffusion Solvers}
\subsection{Active Learning} \label{sub_active_learning}
Let $D=(x, y)$ be the complete input and output training data pair divided into the initially labeled set $L$ and the remaining unlabeled data $U$. We initially train the diffusion surrogate model with the $L$ and evaluate the acquisition function on $U$. $B$ most informative data from $U$ are labeled and appended to $L$ such that $L = L\ \cup\ B$ and $ U = U\ \backslash\ B$. The model is retrained on the labeled data and the process iterates until a performance metric $P$ is reached or a total labeled data size reaches $S$. 

Note that an important consideration for the design of acquisition functions is that the labels (in this case the simulation outputs on the given inputs) of the samples are not available before the samples are selected to be queried for label. 
In another word, 
the acquisition objective cannot utilize the label of the samples. 
We consider three active learning acquisition functions and compare them with random acquisition. Two are representative of uncertainty-based acquisitions, including the use of entropy \cite{joshi2009multi} 
and the estimated loss of the DNN surrogage \cite{huang2021semi}. One is representative of acquisitions considering the diversity of training samples.

\paragraph{Entropy:} Entropy sampling \cite{joshi2009multi} selects instances that the model is most uncertain about as measured by the standard deviation of the 'k' output predictions of the network. We use dropout as Bayesian approximation \cite{gal2016dropout} to generate 'k' different output predictions and evaluate entropy as:

\begin{equation} \label{eqn_entropy}
    Entropy = \frac{1}{Nk}\sum_{j=1}^{N}\sum_{r \in \{1, 2, ..., k\}}(\hat{y}_{rj} - \bar{y}_j)^2
\end{equation}

where, $\hat{y}_{rj}$ is the $j^{th}$ pixel in the $r^{th}$ prediction, $\bar{y}_j$ is the mean prediction of the $j^{th}$ pixel obtained through 'k' forward passes with dropout
and $N$ are the total number of pixels in the lattice. 

\begin{figure}[hbtp]
  \centering
  \includegraphics[width=4.0in]{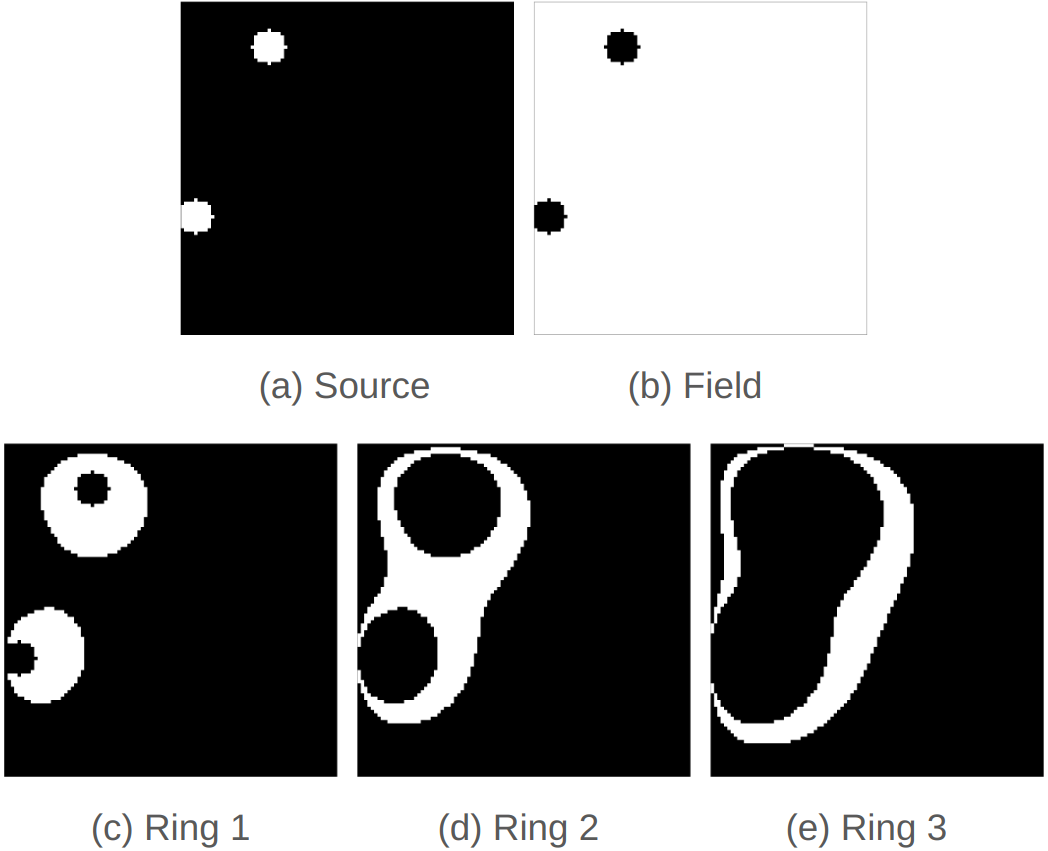}
  \caption{Regions of interest shown with white for a given lattice -- a) Source, b) Field, c) Ring 1, d) Ring 2 and e) Ring 3 for two sources placed randomly in the lattice.}
  \label{fig_regions}
\end{figure}

\paragraph{Temporal Output Discrepancy (TOD):} 
The loss of the DNN surrogate on a new sample is another potential criteria for selecting samples 
a DNN is most uncertain about:
a high-loss sample has the potential to provide stronger signals for the optimization of the DNN, 
whereas a low-loss sample could be redundant to what the DNN is already trained with. 
Intended as a criteria for selecting unlabeled data samples, 
the loss of the DNN on the unlabeled samples cannot be calculated from the actual labels (here the simulation outputs) but must be estimated. 
Following the Temporal Output Discrepancy (TOD) measure \cite{huang2021semi}, we 
estimate the DNN loss on a sample based on the discrepancy of the DNN outputs on the same sample at different learning iterations: 
\begin{equation} \label{eqn_tod}
    D_t^{T}(x) \overset{\mathrm{def}}{=} ||f(x;\theta_{t+T} - f(x;\theta_t)||
\end{equation}
where $D_t^{T}(x)$ is the distance between outputs of the model $f$ with parameter $\theta_{t+T}$ and $\theta_{t}$ evaluated at $(t+T)^{th}$ and $t^{th}$ gradient optimization step. In our experiments, we set $T\ =\ 1$.

To ground the results obtained from TOD, we also use the \textit{actual} loss of a new sample as the acquisition criteria 
owing to the existence of simulation data already generated offline. 
Once we train the DNN surrogate on the labeled data $L$, we calculate the loss of the unlabelled data $U$ with the available simulation outputs and select $B$ examples with the highest losses to add to $L$ for the next training. Note that this is intended to provide a reference for the efficacy of TOD that is reliant on an estimation of the DNN loss on unlabled samples, not an acquisition strategy 
to be used during active learning of DNN surrogates (as true loss is not available on unlabeled new samples).

\paragraph{Parameter Diversity:} 
To consider the diversity of training simulations without looking at the simulated outputs, we measure diversity by the parameters used to generate the simulations. 
We consider six parameters in the input initial condition, including the coordinates (cx and cy) of the two sources ($cx_1, cy_1, cx_2, cy_2$), 
the distance between the sources ($d$), and the intensity of the second source ($Q$)
as the important identifiers ($I$) for the model. Following \cite{sener2017active}, we create a coreset subset from the unlabeled data such that the examples with the largest minimal distance from the labeled data are selected.

\begin{equation} \label{eqn_diversity}
    dist = max_{m \in U}\ min_{n \in L}\bigtriangleup(I_m, I_n)
\end{equation}
We select $B$ examples from $U$ that have maximum calculated parameter diversity distance $dist$.

\subsection{Evaluation Metrics}
\paragraph{Weighted-MAE} 
We assess the DNN surrogate's performance by calculating its loss, \textit{i.e.,} its weighted mean absolute error, as defined in Equation ~\ref{eqn_loss}. After each acquisition round, the metric is computed on the test data for all active learning acquisition functions. 

\paragraph{Region-of-Interest MAE} 
As mentioned earlier, the sources occupy 2\% of the total lattice space where the diffusion pattern emanate from the source and die out quickly in the field. 
To better understand the effect of active learning on the accuracy of DNN surrogates, 
we further 
examine the MAE in different regions of interest on the lattice as illustrated in Figure~\ref{fig_regions}: 
the regions of 
"Source" and "Field" are determined based on the initial inputs to the simulations; the regions of "RINGS 1," "RINGS 2," and "RINGS 3" are defined based on the ground-truth simulation outputs using 
pixel values in the ranges [0.2, 1.0], [0.1, 0.2], and [0.05, 0.1], respectively -- representing increasing distances to the sources.

\section{Experiments}

We split the two-source data described in section \ref{sub_data}. into 16,000 training data, 4,000 validation data, and 4,000 test data. We trained the U-Net and CNN autoencoder structure defined in section \ref{sub_model}. We divided the 16000 training data into an initial 1000 labelled data (samples with simulation outputs available) and a 15000 unlabelled pool (samples without simulation outputs). We performed active learning with the acquisition functions defined in section \ref{sub_active_learning} for CNN autoencoder and exclude entropy acquisition function for U-Net due to large computational demand for $'N'$ forward passes. At each acquisition round, 
new unlabled samples are selected to query for simulation outputs, and 
the DNN surrogates were retrained with the current labeled dataset (input-output simulationp pairs). 
This was repeated until all the 15000 unlabelled samples are labeled. In each active learning round, the DNN surrogate was trained for 500 epochs and saved based on the best validation loss. For the entropy acquisition function, in CNN autoencoder, 40\% dropouts were added after batch normalization of the first and second convolutional layers. The code repository can be found in \cite{pbgithub}.


\begin{figure}[hptb]
  \centering
  \includegraphics[width=\linewidth]{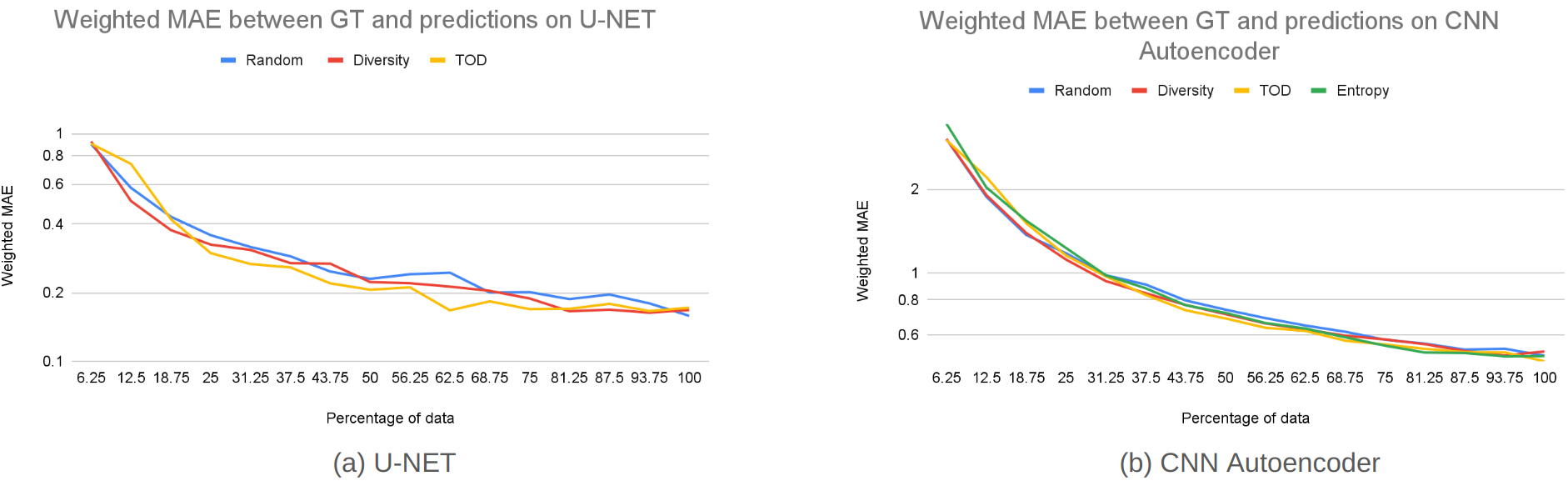}
  \caption{Comparison of test weighted-MAE on a) U-Net architecture and  b) CNN autoencoder, in log scale, between the ground truth and prediction from the DNN surrogates across different acquisition functions as data acquisition proceeds.}
  \label{fig_acq_compare}
  \includegraphics[width=\linewidth]{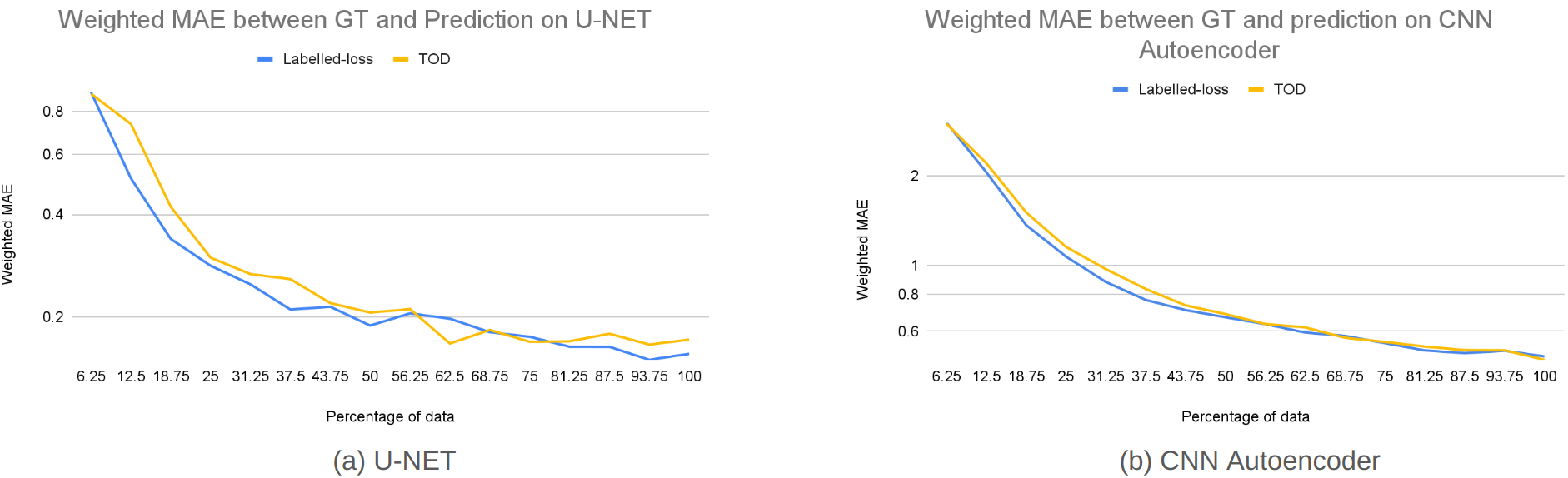}
  \caption{Comparison of test weighted-MAE on a) U-Net architecture and  b) CNN autoencoder, in log scale, between the ground truth and prediction from the DNN surrogates between data acquisitions using TOD and labelled-loss as data acquisition proceeds.}
  \label{fig_todloss_compare}
\end{figure} 
\subsection{Comparison of Active Learning Acquisition Functions} 
\label{sub_al_compare} 


Figure~\ref{fig_acq_compare}(a) compares weighted-MAE between the ground truth and DNN surrogate predictions for different acquisition functions at different percentages of labeled data, 
evaluated against random acquisitions for the U-Net architecture. 
As shown, 
actively selecting simulation data for DNN surrogates
training, 
especially by considering the estimated loss of the surrogate (TOD), 
consistently improves the DNN performance compared to random acquisition 
using the same amount of labeled training data. 
Note that TOD, even though reliant on an estimate of the DNN loss without the simulation output as labels, 
acquires a performance on par with using the actual DNN loss calculated from labels as shown in Figure~\ref{fig_todloss_compare}(a). 
In comparison, 
 diversity-based selection of training samples 
demonstrated limited benefits 
with 
a performance similar to or marginally better than random acquisition. 

\begin{figure*}[!t]
  \centering
  \includegraphics[width=\linewidth]{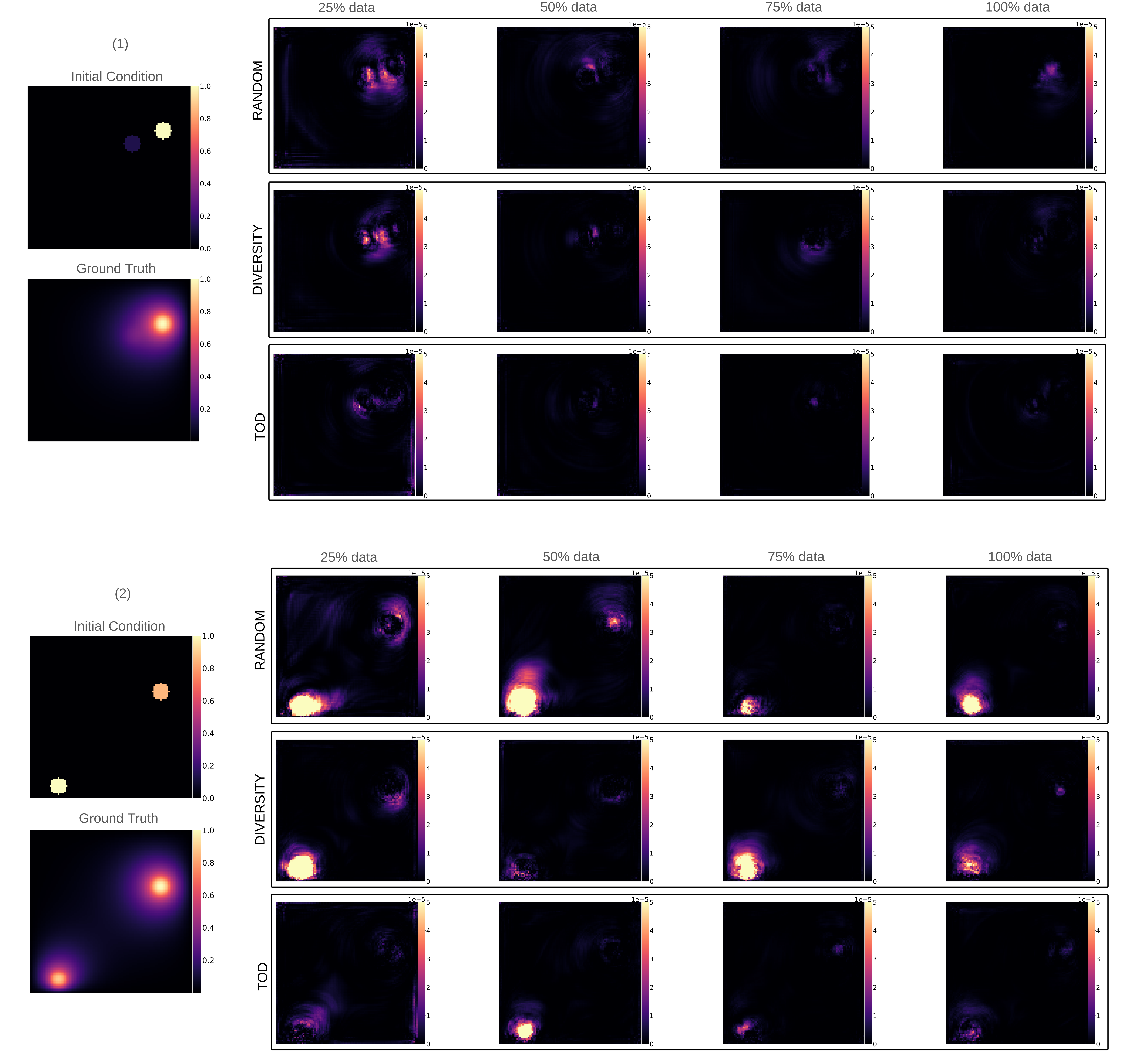}
  \caption{Comparison of absolute error maps between DNN predictions and the ground truth using between random (top row), diversity (middle row), and TOD (bottom row) acquisition function at different percentage 
  of labelled data on U-Net architecture for two example lattice (1) and (2)}
  \label{fig_unet_error_diffpercent}
\end{figure*}

Table ~\ref{table_unet_cnn} lists the 
weighted MAE between the simulated ground truth and predictions on the test data for U-Net across quarterly percentages of labeled data, computed for the different regions of interest as described in Figure~\ref{fig_regions}. 
The results show that, 
at the same amount of labeled training data throughout the active learning process, 
the careful selection of training simulations achieved a test performance improvement across all acquisition strategies (except when all data are used): this improvement was more evident when the size of the labeled data was small. 
Additionally, these improvements were more pronounced at the regions proximal to the source (SRC, RING1) where the errors were the highest. 
This observation can be better appreciated in visual examples of the 
absolute errors 
(Figure~\ref{fig_unet_error_diffpercent}) 
between DNN surrogate predictions and ground-truth simulation outputs obtained on the U-Net architecture. 
As expected, 
across the three rows, 
there is a  gradual decrease in error as the training data increases. 
Notably, however, TOD consistently exhibits lower errors compared to random training, 
especially in regions close to the sources. 

\begin{figure*}[!h]
  \centering
  \includegraphics[width=\linewidth]{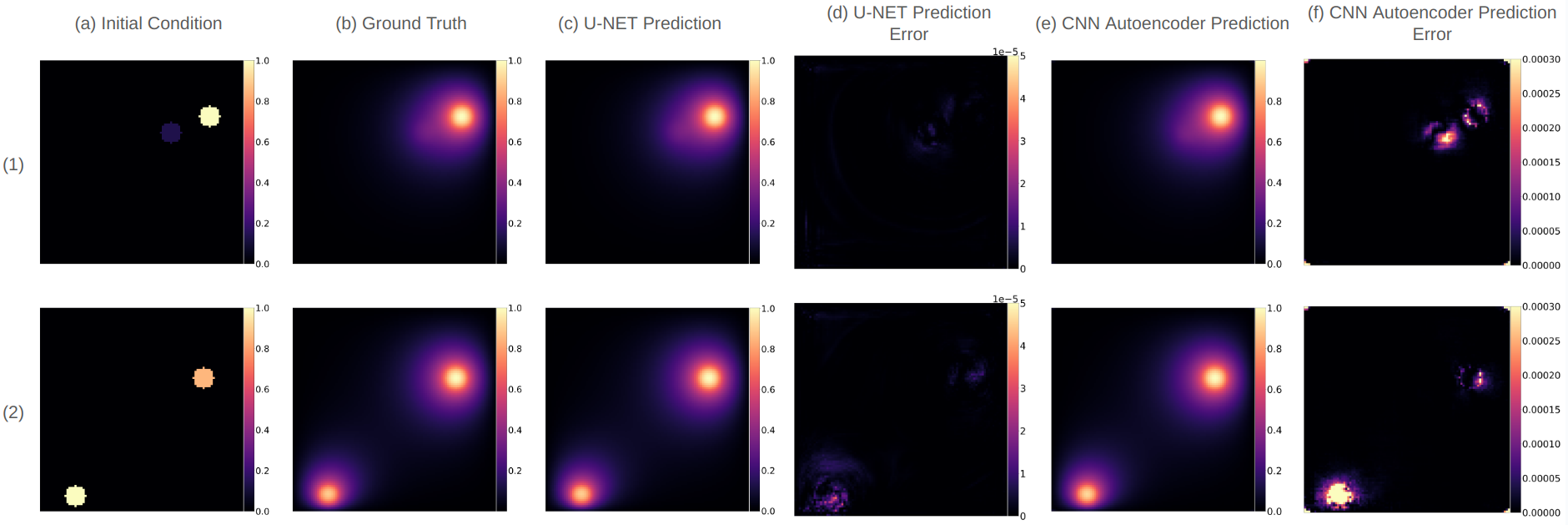}
  \caption{(a) Initial condition, (b) steady-state solution (ground truth), prediction of (c) U-Net and (e) CNN autoencoder surrogate, and prediction error of (d) U-Net and (f) CNN autoencoder surrogate, trained on 100\% data, for two examples of two source simulation data where sources of 5-pixel radius are placed randomly within a 100 x 100 lattice. The scale for the error color bar for UNET and CNN autoencoder are different.}
  \label{fig_unet_cnn_prediction}
\end{figure*}

\begin{figure*}[!t]
  \centering
  \includegraphics[width=\linewidth]{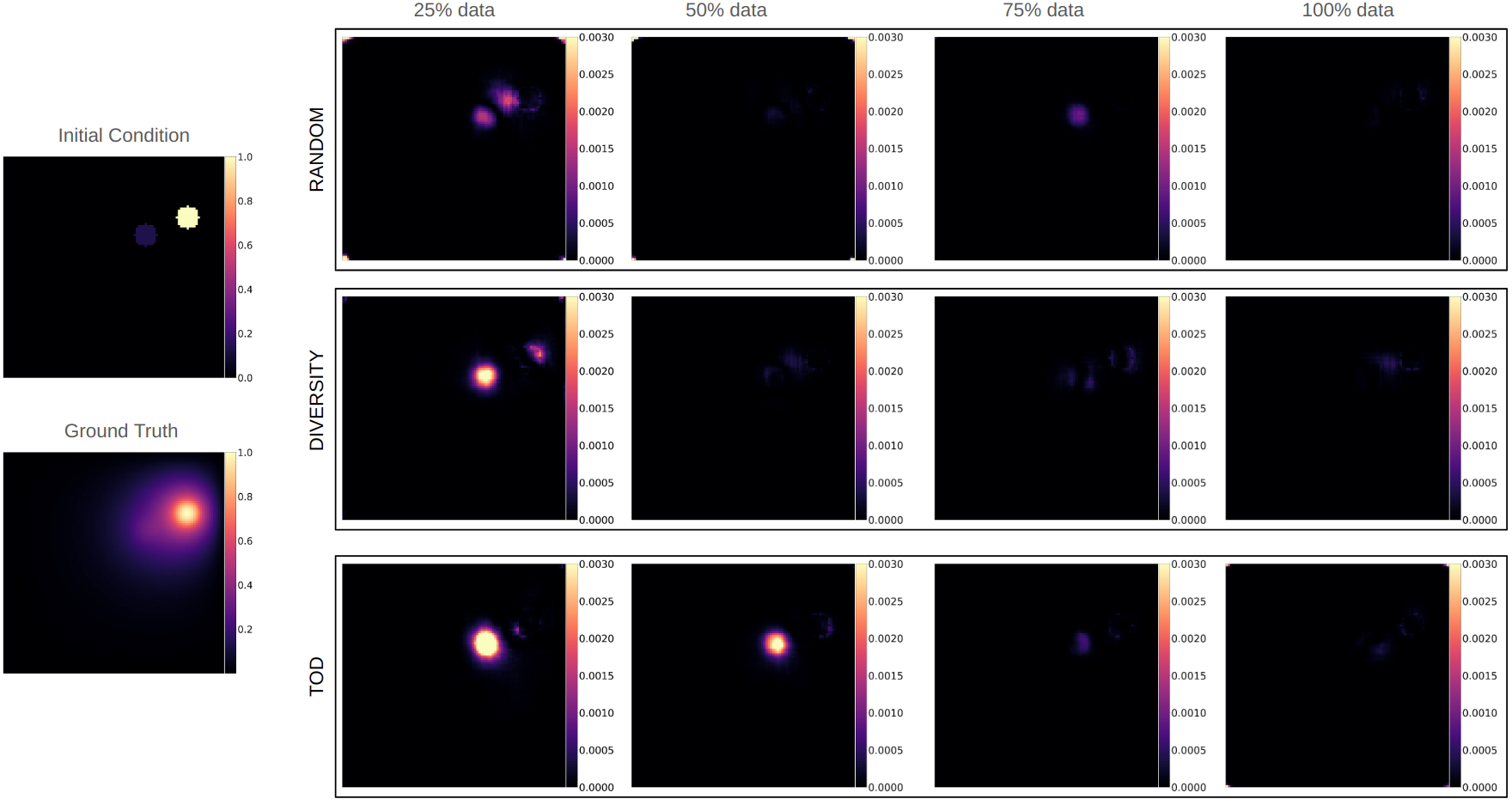}
  \caption{Comparison of absolute error maps between DNN predictions and the ground truth using between random (top row), diversity (middle row), and TOD (bottom row) acquisition function at different percentage 
  of labelled data on CNN autoencoder. The scale for the error color bar is different from U-Net in Figure~\ref{fig_unet_error_diffpercent}}
  \label{fig_cnn_error_diffpercent}
\end{figure*}
\begin{table*}[h!]
\centering
\resizebox{\textwidth}{!}{
\begin{tabular}{lllllcllrrrrc}
\cline{2-5} \cline{9-12}
\multicolumn{1}{l|}{}         & \multicolumn{4}{c|}{Test Loss for U-Net}                                                                                                               & \multicolumn{1}{l}{}       &                       & \multicolumn{1}{l|}{}        & \multicolumn{4}{c|}{Test Loss for CNN Autoencoder}                                                                                                       & \multicolumn{1}{l}{}       \\ \cline{2-5} \cline{9-12}
\multicolumn{1}{l|}{}         & \multicolumn{4}{c|}{DATA SIZES}                                                                                                                        & \multicolumn{1}{l}{}       &                       & \multicolumn{1}{l|}{}        & \multicolumn{4}{c|}{DATA SIZES}                                                                                                                          & \multicolumn{1}{l}{}       \\ \cline{2-5} \cline{9-12}
\multicolumn{1}{l|}{}         & \multicolumn{1}{l|}{25\%}      & \multicolumn{1}{l|}{50\%}      & \multicolumn{1}{l|}{75\%}      & \multicolumn{1}{l|}{100\%}                          & \multicolumn{1}{l}{}       &                       & \multicolumn{1}{l|}{}        & \multicolumn{1}{l|}{25\%}       & \multicolumn{1}{l|}{50\%}       & \multicolumn{1}{l|}{75\%}      & \multicolumn{1}{l|}{100\%}                          & \multicolumn{1}{l}{}       \\ \cline{1-6} \cline{8-13} 
\multicolumn{1}{|l|}{Random}  & 0.35784                        & 0.23011                        & 0.20112                        & \multicolumn{1}{l|}{0.15839}                        & \multicolumn{1}{c|}{}      & \multicolumn{1}{l|}{} & \multicolumn{1}{l|}{Random}  & 1.17626                         & 0.73720                         & 0.57466                        & \multicolumn{1}{r|}{0.50335}                        & \multicolumn{1}{c|}{}      \\ \cline{1-1} \cline{8-8}
\multicolumn{1}{|l|}{Diverse} & {\color[HTML]{FE0000} 0.32529} & {\color[HTML]{FE0000} 0.22275} & {\color[HTML]{FE0000} 0.18868} & \multicolumn{1}{l|}{0.16786}                        & \multicolumn{1}{c|}{}      & \multicolumn{1}{l|}{} & \multicolumn{1}{l|}{Diverse} & {\color[HTML]{FE0000} 1.11622}  & {\color[HTML]{FE0000} 0.71032}  & 0.57625                        & \multicolumn{1}{r|}{0.52171}                        & \multicolumn{1}{c|}{}      \\ \cline{1-1} \cline{8-8}
\multicolumn{1}{|l|}{TOD}     & {\color[HTML]{FE0000} 0.29847} & {\color[HTML]{FE0000} 0.20616} & {\color[HTML]{FE0000} 0.16936} & \multicolumn{1}{l|}{0.16616}                        & \multicolumn{1}{c|}{ALL}   & \multicolumn{1}{l|}{} & \multicolumn{1}{l|}{TOD}     & {\color[HTML]{FE0000} 1.15445}  & {\color[HTML]{FE0000} 0.68581}  & {\color[HTML]{FE0000} 0.55251} & \multicolumn{1}{r|}{{\color[HTML]{FE0000} 0.48098}} & \multicolumn{1}{c|}{}      \\ \cline{1-1} \cline{6-6} \cline{8-8}
                              & \multicolumn{1}{r}{}           & \multicolumn{1}{r}{}           & \multicolumn{1}{r}{}           & \multicolumn{1}{r}{}                                & \multicolumn{1}{l}{}       & \multicolumn{1}{l|}{} & \multicolumn{1}{l|}{Entropy} & 1.22860                         & {\color[HTML]{FE0000} 0.71770}  & {\color[HTML]{FE0000} 0.54734} & \multicolumn{1}{r|}{{\color[HTML]{FE0000} 0.50217}} & \multicolumn{1}{c|}{ALL}   \\ \cline{8-8} \cline{13-13} 
                              &                                &                                &                                &                                                     & \multicolumn{1}{l}{}       &                       &                              & \multicolumn{1}{l}{}            & \multicolumn{1}{l}{}            & \multicolumn{1}{l}{}           & \multicolumn{1}{l}{}                                & \multicolumn{1}{l}{}       \\ \cline{1-1} \cline{6-6} \cline{8-8} \cline{13-13} 
\multicolumn{1}{|l|}{Random}  & 4.92366                        & 3.13306                        & 2.75548                        & \multicolumn{1}{l|}{2.19644}                        & \multicolumn{1}{c|}{}      & \multicolumn{1}{l|}{} & \multicolumn{1}{l|}{Random}  & 18.46547                        & 11.72694                        & 8.68628                        & \multicolumn{1}{r|}{7.60275}                        & \multicolumn{1}{c|}{}      \\ \cline{1-1} \cline{8-8}
\multicolumn{1}{|l|}{Diverse} & {\color[HTML]{FE0000} 4.41641} & {\color[HTML]{FE0000} 2.96436} & {\color[HTML]{FE0000} 2.52532} & \multicolumn{1}{l|}{{\color[HTML]{FE0000} 2.17466}} & \multicolumn{1}{c|}{}      & \multicolumn{1}{l|}{} & \multicolumn{1}{l|}{Diverse} & {\color[HTML]{FE0000} 18.03126} & {\color[HTML]{FE0000} 11.03576} & {\color[HTML]{FE0000} 8.59056} & \multicolumn{1}{r|}{7.99048}                        & \multicolumn{1}{c|}{}      \\ \cline{1-1} \cline{8-8}
\multicolumn{1}{|l|}{TOD}     & {\color[HTML]{FE0000} 3.58799} & {\color[HTML]{FE0000} 2.91486} & {\color[HTML]{FE0000} 2.07431} & \multicolumn{1}{l|}{2.33184}                        & \multicolumn{1}{c|}{SRC}   & \multicolumn{1}{l|}{} & \multicolumn{1}{l|}{TOD}     & {\color[HTML]{FE0000} 18.29991} & {\color[HTML]{FE0000} 10.79857} & {\color[HTML]{FE0000} 8.45668} & \multicolumn{1}{r|}{{\color[HTML]{FE0000} 7.16640}} & \multicolumn{1}{c|}{}      \\ \cline{1-1} \cline{6-6} \cline{8-8}
                              & \multicolumn{1}{r}{}           & \multicolumn{1}{r}{}           & \multicolumn{1}{r}{}           & \multicolumn{1}{r}{}                                & \multicolumn{1}{l}{}       & \multicolumn{1}{l|}{} & \multicolumn{1}{l|}{Entropy} & 18.48527                        & {\color[HTML]{FE0000} 11.08221} & {\color[HTML]{FE0000} 8.19734} & \multicolumn{1}{r|}{7.73362}                        & \multicolumn{1}{c|}{SRC}   \\ \cline{8-8} \cline{13-13} 
                              &                                &                                &                                &                                                     & \multicolumn{1}{l}{}       &                       &                              & \multicolumn{1}{l}{}            & \multicolumn{1}{l}{}            & \multicolumn{1}{l}{}           & \multicolumn{1}{l}{}                                & \multicolumn{1}{l}{}       \\ \cline{1-1} \cline{6-6} \cline{8-8} \cline{13-13} 
\multicolumn{1}{|l|}{Random}  & 1.48035                        & 0.94344                        & 0.84136                        & \multicolumn{1}{l|}{0.63358}                        & \multicolumn{1}{c|}{}      & \multicolumn{1}{l|}{} & \multicolumn{1}{l|}{Random}  & 5.33343                         & 3.30900                         & 2.63469                        & \multicolumn{1}{r|}{2.31095}                        & \multicolumn{1}{c|}{}      \\ \cline{1-1} \cline{8-8}
\multicolumn{1}{|l|}{Diverse} & {\color[HTML]{FE0000} 1.41882} & {\color[HTML]{FE0000} 0.92979} & {\color[HTML]{FE0000} 0.79940} & \multicolumn{1}{l|}{0.69533}                        & \multicolumn{1}{c|}{}      & \multicolumn{1}{l|}{} & \multicolumn{1}{l|}{Diverse} & {\color[HTML]{FE0000} 5.01338}  & {\color[HTML]{FE0000} 3.22305}  & 2.65821                        & \multicolumn{1}{r|}{2.36826}                        & \multicolumn{1}{c|}{}      \\ \cline{1-1} \cline{8-8}
\multicolumn{1}{|l|}{TOD}     & {\color[HTML]{FE0000} 1.15558} & {\color[HTML]{FE0000} 0.82894} & {\color[HTML]{FE0000} 0.69378} & \multicolumn{1}{l|}{0.67992}                        & \multicolumn{1}{c|}{RING1} & \multicolumn{1}{l|}{} & \multicolumn{1}{l|}{TOD}     & {\color[HTML]{FE0000} 5.02481}  & {\color[HTML]{FE0000} 3.06267}  & {\color[HTML]{FE0000} 2.51028} & \multicolumn{1}{r|}{{\color[HTML]{FE0000} 2.19476}} & \multicolumn{1}{c|}{}      \\ \cline{1-1} \cline{6-6} \cline{8-8}
                              & \multicolumn{1}{r}{}           & \multicolumn{1}{r}{}           & \multicolumn{1}{r}{}           & \multicolumn{1}{r}{}                                & \multicolumn{1}{l}{}       & \multicolumn{1}{l|}{} & \multicolumn{1}{l|}{Entropy} & 5.52691                         & {\color[HTML]{FE0000} 3.20892}  & {\color[HTML]{FE0000} 2.48799} & \multicolumn{1}{r|}{{\color[HTML]{FE0000} 2.27788}} & \multicolumn{1}{c|}{RING1} \\ \cline{8-8} \cline{13-13} 
                              &                                &                                &                                &                                                     & \multicolumn{1}{l}{}       &                       &                              & \multicolumn{1}{l}{}            & \multicolumn{1}{l}{}            & \multicolumn{1}{l}{}           & \multicolumn{1}{l}{}                                & \multicolumn{1}{l}{}       \\ \cline{1-1} \cline{6-6} \cline{8-8} \cline{13-13} 
\multicolumn{1}{|l|}{Random}  & 0.34664                        & 0.23742                        & 0.19985                        & \multicolumn{1}{l|}{0.17337}                        & \multicolumn{1}{c|}{}      & \multicolumn{1}{l|}{} & \multicolumn{1}{l|}{Random}  & 0.90847                         & 0.57265                         & 0.45602                        & \multicolumn{1}{r|}{0.39152}                        & \multicolumn{1}{c|}{}      \\ \cline{1-1} \cline{8-8}
\multicolumn{1}{|l|}{Diverse} & {\color[HTML]{FE0000} 0.30417} & {\color[HTML]{FE0000} 0.23389} & {\color[HTML]{FE0000} 0.18549} & \multicolumn{1}{l|}{{\color[HTML]{FE0000} 0.17087}} & \multicolumn{1}{c|}{}      & \multicolumn{1}{l|}{} & \multicolumn{1}{l|}{Diverse} & {\color[HTML]{FE0000} 0.86313}  & {\color[HTML]{FE0000} 0.54783}  & {\color[HTML]{FE0000} 0.44496} & \multicolumn{1}{r|}{0.41255}                        & \multicolumn{1}{c|}{}      \\ \cline{1-1} \cline{8-8}
\multicolumn{1}{|l|}{TOD}     & 0.37816                        & {\color[HTML]{FE0000} 0.21057} & {\color[HTML]{FE0000} 0.19488} & \multicolumn{1}{l|}{{\color[HTML]{FE0000} 0.17043}} & \multicolumn{1}{l|}{RING2} & \multicolumn{1}{l|}{} & \multicolumn{1}{l|}{TOD}     & 1.00724                         & {\color[HTML]{FE0000} 0.54755}  & {\color[HTML]{FE0000} 0.43630} & \multicolumn{1}{r|}{{\color[HTML]{FE0000} 0.38701}} & \multicolumn{1}{c|}{}      \\ \cline{1-1} \cline{6-6} \cline{8-8}
                              &                                &                                &                                &                                                     &                            & \multicolumn{1}{l|}{} & \multicolumn{1}{l|}{Entropy} & 1.03330                         & 0.59488                         & {\color[HTML]{FE0000} 0.44425} & \multicolumn{1}{r|}{0.39372}                        & \multicolumn{1}{l|}{RING2} \\ \cline{8-8} \cline{13-13} 
                              &                                &                                &                                &                                                     & \multicolumn{1}{l}{}       &                       &                              & \multicolumn{1}{l}{}            & \multicolumn{1}{l}{}            & \multicolumn{1}{l}{}           & \multicolumn{1}{l}{}                                &                            \\ \cline{1-1} \cline{6-6} \cline{8-8} \cline{13-13} 
\multicolumn{1}{|l|}{Random}  & 0.14096                        & 0.09395                        & 0.07416                        & \multicolumn{1}{l|}{0.06339}                        & \multicolumn{1}{c|}{}      & \multicolumn{1}{l|}{} & \multicolumn{1}{l|}{Random}  & 0.27176                         & 0.17411                         & 0.13709                        & \multicolumn{1}{r|}{0.11720}                        & \multicolumn{1}{c|}{}      \\ \cline{1-1} \cline{8-8}
\multicolumn{1}{|l|}{Diverse} & {\color[HTML]{FE0000} 0.11276} & {\color[HTML]{FE0000} 0.08636} & {\color[HTML]{FE0000} 0.06804} & \multicolumn{1}{l|}{0.07206}                        & \multicolumn{1}{c|}{}      & \multicolumn{1}{l|}{} & \multicolumn{1}{l|}{Diverse} & {\color[HTML]{FE0000} 0.25081}  & {\color[HTML]{FE0000} 0.16577}  & 0.13732                        & \multicolumn{1}{r|}{0.12434}                        & \multicolumn{1}{c|}{}      \\ \cline{1-1} \cline{8-8}
\multicolumn{1}{|l|}{TOD}     & 0.15077                        & {\color[HTML]{FE0000} 0.08337} & 0.07653                        & \multicolumn{1}{l|}{0.06613}                        & \multicolumn{1}{l|}{RING3} & \multicolumn{1}{l|}{} & \multicolumn{1}{l|}{TOD}     & 0.30949                         & {\color[HTML]{FE0000} 0.17163}  & {\color[HTML]{FE0000} 0.13658} & \multicolumn{1}{r|}{0.11916}                        & \multicolumn{1}{c|}{}      \\ \cline{1-1} \cline{6-6} \cline{8-8}
                              &                                &                                &                                &                                                     &                            & \multicolumn{1}{l|}{} & \multicolumn{1}{l|}{Entropy} & 0.31644                         & 0.18589                         & {\color[HTML]{FE0000} 0.13581} & \multicolumn{1}{r|}{0.12081}                        & \multicolumn{1}{c|}{RING3} \\ \cline{8-8} \cline{13-13} 
                              &                                &                                &                                &                                                     &                            &                       &                              & \multicolumn{1}{l}{}            & \multicolumn{1}{l}{}            & \multicolumn{1}{l}{}           & \multicolumn{1}{l}{}                                &                            \\ \cline{1-1} \cline{6-6} \cline{8-8} \cline{13-13} 
\multicolumn{1}{|l|}{Random}  & 0.28301                        & 0.18254                        & 0.15927                        & \multicolumn{1}{l|}{0.12500}                        & \multicolumn{1}{c|}{}      & \multicolumn{1}{l|}{} & \multicolumn{1}{l|}{Random}  & 0.89296                         & 0.55713                         & 0.44174                        & \multicolumn{1}{r|}{0.38700}                        & \multicolumn{1}{c|}{}      \\ \cline{1-1} \cline{8-8}
\multicolumn{1}{|l|}{Diverse} & {\color[HTML]{FE0000} 0.25827} & {\color[HTML]{FE0000} 0.17783} & {\color[HTML]{FE0000} 0.15039} & \multicolumn{1}{l|}{{\color[HTML]{FE0000} 0.13497}} & \multicolumn{1}{c|}{}      & \multicolumn{1}{l|}{} & \multicolumn{1}{l|}{Diverse} & {\color[HTML]{FE0000} 0.83903}  & {\color[HTML]{FE0000} 0.54113}  & 0.44492                        & \multicolumn{1}{r|}{0.39932}                        & \multicolumn{1}{c|}{}      \\ \cline{1-1} \cline{8-8}
\multicolumn{1}{|l|}{TOD}     & {\color[HTML]{FE0000} 0.24457} & {\color[HTML]{FE0000} 0.16177} & {\color[HTML]{FE0000} 0.13815} & \multicolumn{1}{l|}{0.13069}                        & \multicolumn{1}{c|}{FIELD} & \multicolumn{1}{l|}{} & \multicolumn{1}{l|}{TOD}     & {\color[HTML]{FE0000} 0.87351}  & {\color[HTML]{FE0000} 0.52009}  & {\color[HTML]{FE0000} 0.42297} & \multicolumn{1}{r|}{{\color[HTML]{FE0000} 0.37143}} & \multicolumn{1}{c|}{}      \\ \cline{1-1} \cline{6-6} \cline{8-8}
                              &                                &                                &                                &                                                     & \multicolumn{1}{l}{}       & \multicolumn{1}{l|}{} & \multicolumn{1}{l|}{Entropy} & 0.94582                         & {\color[HTML]{FE0000} 0.54786}  & {\color[HTML]{FE0000} 0.42197} & \multicolumn{1}{r|}{{\color[HTML]{FE0000} 0.38366}} & \multicolumn{1}{c|}{FIELD} \\ \cline{8-8} \cline{13-13} 
\end{tabular}}
\caption{Comparison of test MAE between the prediction of the DNN surrogates and ground-truth simulations using U-Net (left) and CNN autoencoder (right) network for different percentages of labelled data across different acquisition functions. The metrics are measured for different lattice regions of interest as depicted in Figure \ref{fig_regions}. Red indicate measures better or similar to the random acquisition.}
\label{table_unet_cnn}
\end{table*}

Overall, 
TOD was able to use a smaller amount of data ($\sim$50\%) to achieve the performance random acquisition can achieve with a higher amount of data ($\sim$75\%). 
Note that we do not expect significant differences between different acquisition strategies when 100\% data are used.


\subsection{Effect of Architecture on Active Learning} \label{sub_eff_arch}


Figure \ref{fig_unet_cnn_prediction} 
first compares the results obtained by U-Net and CNN autoencoder when 100\% data were used, which shows that CNN autoencoder was suboptimal as a surrogate architecture compared to the U-Net for the diffusion equations. 

Figure~\ref{fig_acq_compare}(b) compares the weighted MAE for the CNN autoencoder at different percentages of labeled data across different acquisition functions. Compared with Figure~\ref{fig_acq_compare}(a), it is apparent that despite using identical acquisition functions and datasets, the performance improvements due to active learning is more pronounced in U-Net architecture compared to CNN autoencoder: the performance gain due to active learning for the CNN autoencoder is marginal at best with random training. This distinction is also evident in the quantitative data presented in Table ~\ref{table_unet_cnn}. Similarly, as shown in the the visual examples of the absolute error maps between ground-truth simulations and surrogate predictions presented in Figure~\ref{fig_cnn_error_diffpercent}, in contrast to our previous observations on the U-Net, the errors among the acquisition functions remain consistently similar.

These results suggest an interesting and important finding: 
for active selection of training simulations to play a role in the construction of DNN surrogates, 
it is important to first identify an appropriate if not optimal DNN architecture for the surrogate, as 
the choices of architectures have a significant effect on the relative ranking of various acquisition function and the benefit they can deliver. Interestingly, this result is consistent with those reported systematic evaluations of general DAL methods \cite{beck2021effective, munjal2022towards}.


\section{Discussion}
In this paper, we investigate the feasibility of integrating active learning in the training of a DNN surrogate for the diffusion solver. Our findings highlight two key observations. Firstly, training a DNN surrogate with intelligently selected simulations has the potential to reduce the requirement on the generation of expensive simulations and improve the performance of the DNN surrogates. 
Specifically, acquisition strategies 
focused on the predicted loss of the DNN surrogates on new samples may be the most promising for training smart DNN surrogates. 
This encourages the use of active learning in training surrogate models with less but informative data rather than a pre-annotated dataset. Secondly, the choice of network itself significantly impacts the benefit derived from active learning: with the same data and acquisition strategies, 
to what extent the active learning 
improves the DNN surrogate training largely depends on the underlying choice of DNN architecture -- CNN autoencoder versus U-Net in this case. 
This suggests that, 
to develop a HPC infrastructure that support the construction of Smart Surrogate with active learning, 
an additional component that needs to be supported by the infrastructure may be the 
optimization of the DNN surrogate architecture prior to active learning.


As a first proof-of-concept feasibility study, future works can be improved along the following fronts. 

\paragraph{Diversification of Applications and Dataset Scales} While our current work is focused on a use case of diffusion solver surrogate with two sources randomly placed on a $100^2$ lattice, we plan on incorporating a larger simulation set. This set will feature a variable number of sources on a larger lattice, capturing inherent randomness in real world simulations. Furthermore, we recognize the importance to broaden our application scope, transcending the confines of our current focus to encompass a diverse array of use cases. 

\paragraph{Broader Spectrums of Acquisition Functions}  Our study currently incorporates three distinctive acquisition functions, serving as the foundation for exploring the integration of active learning into surrogate training. Owing to the ever-developing field of active learning field, our future plans involve the inclusion of additional acquisition functions that leverage evidential uncertainty and hybrid strategies, combining the benefit of uncertainty and diversity. 

\paragraph{Expansive Exploration of Architectural Impact} The interplay between active learning performance and architecture 
as observed in this study provides important insight into 
the importance of optimal architectural design and training data selection in building DNN surrogates. Future works will 
entail an empirical study over a larger architecture space and even branching out to fields like architecture optimization and network architecture search. 

\paragraph{Transition to Smart Surrogates: On-the-Fly  Simulations Steered by Active Learning} This study, although based on simulation data generated offline, provide important insights into the feasibility and key design elements for active learning of DNN surrogates. 
As a pivotal shift in our research trajectory, we are moving from offline emulation to online scenarios where HPC simulations will be steered and executed on the fly by active learning. 
While the methodological framework presented in this study will generally apply with minimal modifications, substantial efforts will be needed to establish the HPC infrastructure that can support on-the-fly allocation of different HPC resources in between the execution of high-performance simulations, DNN training, and decision-making of data acquisitions -- an exciting next step of the current study.

\section{Conclusion}
We present an investigative study that underscores the benefit of utilizing active learning in training a diffusion solver surrogate model. We experimentally show that for certain acquisition functions, active learning with fewer data (<50\% data) shows promise in improving over training randomly with larger data size (>75\% data). This provides a strong foundation for the next steps to build up the HPC infrastructure of \textit{Smart Surrogates} where training simulations are generated on the fly as steered by active learning, potentially on a DNN architecture optimized for the scientific simulations at hand.

\section{Acknowledgements}

This work is supported by National Science Foundation funding NSF OAC-2212548 and NSF OAC-2212550.

\bibliographystyle{unsrt}  
\bibliography{reference}

\end{document}